\documentclass[10pt,twocolumn,letterpaper]{article}

\usepackage{cvpr}              
\usepackage{multirow}

\usepackage[dvipsnames]{xcolor}

\definecolor{cvprblue}{rgb}{0.21,0.49,0.74}
\usepackage[pagebackref,breaklinks,colorlinks,citecolor=cvprblue]{hyperref}

\def\paperID{1218} 
\def\confName{CVPR}
\def\confYear{2024}

\title{Self-Supervised Contrastive Learning for Multi-Label Images}

\author{Jiale Chen\\
College of Computer Science and Software Engineering, Hohai University \\
Nanjing, 210024, China\\
{\tt\small jialechen@hhu.edu.cn}
\and
}

\begin{document}
\maketitle
\begin{abstract}


Self-supervised learning (SSL) has demonstrated its effectiveness in learning representations through comparison methods that align with human intuition. However, mainstream SSL methods heavily rely on high body datasets with single label, such as ImageNet, resulting in intolerable pre-training overhead. Besides, more general multi-label images are frequently overlooked in SSL, despite their potential for richer semantic information and broader applicability in downstream scenarios. Therefore, we tailor the mainstream SSL approach to guarantee excellent representation learning capabilities using fewer multi-label images. Firstly, we propose a block-wise augmentation module aimed at extracting additional potential positive view pairs from multi-label images. Subsequently, an image-aware contrastive loss is devised to establish connections between these views, thereby facilitating the extraction of semantically consistent representations. Comprehensive linear fine-tuning and transfer learning validate the competitiveness of our approach despite challenging sample quality and quantity.


\end{abstract}    
\section{Introduction}
\label{sec:intro}
Self-supervised learning (SSL) has emerged as a compelling approach in the field of computer vision, enabling the acquisition of generalized representations without manual annotation~\cite{survey1,survey2,survey3}. In recent years, substantial advancements have been realized in this field, with optimal self-supervised learning networks now demonstrating performance that competes with supervised learning, particularly in  single-label image classification~\cite{simclr,moco1}.  

To achieve the goal of acquiring  refined backbone weights that can quickly adapted to a variety of downstream tasks without manual labeling, contrastive learning (CL) algorithms have become a widely adopted effective method for  self-supervised pre-training. These algorithms impart  semantically meaningful and coherent visual representations by comparing positive and negative sample pairs, aligning  with human intuition~\cite{simclr,moco1,moco2,BYOL,swav}.

Although  popular self-supervised contrastive learning models have been inherently concise, they consistently relied on long-term training on large datasets, such as ImageNet~\cite{imagenet}. This practice has typically resulted in intolerable computational time and resource overhead. Consequently, contemporary  research has shifted towards utilizing  more compact datasets or datasets explicitly tailored to the requirements of the downstream tasks ~\cite{remote,ship,leaf,rethink,does}. However, these studies all focused  on single-label scenarios rather than more general multi-label cases~\cite{knowledge}.

Statistically, although the classical multi-label dataset COCO2017~\cite{coco} has a larger average image size ($578\times484$) than ImageNet ($473\times406$), the $\approx110k$ samples are only one-tenth of ImageNet. This falls significantly short of the $250k$ samples proposed by Cole~\etal~\cite{does} to balance the self-supervision effect with the time for pre-training. Moreover, the inherent  multi-target coexistence within COCO2017 makes each sample contains multiple potential  targets, which brings significant  challenges to contrastive learning compared to the single-target-centered ImageNet.  Consider that unlabeled images are cheap to acquire and most of them contain multiple semantics rather than single-target central. Therefore, overcoming the challenges of self-supervised learning with multi-label images is an inevitable trend~\cite{knowledge}.

\begin{figure}[t]
  \centering
   \includegraphics[width=1\linewidth]{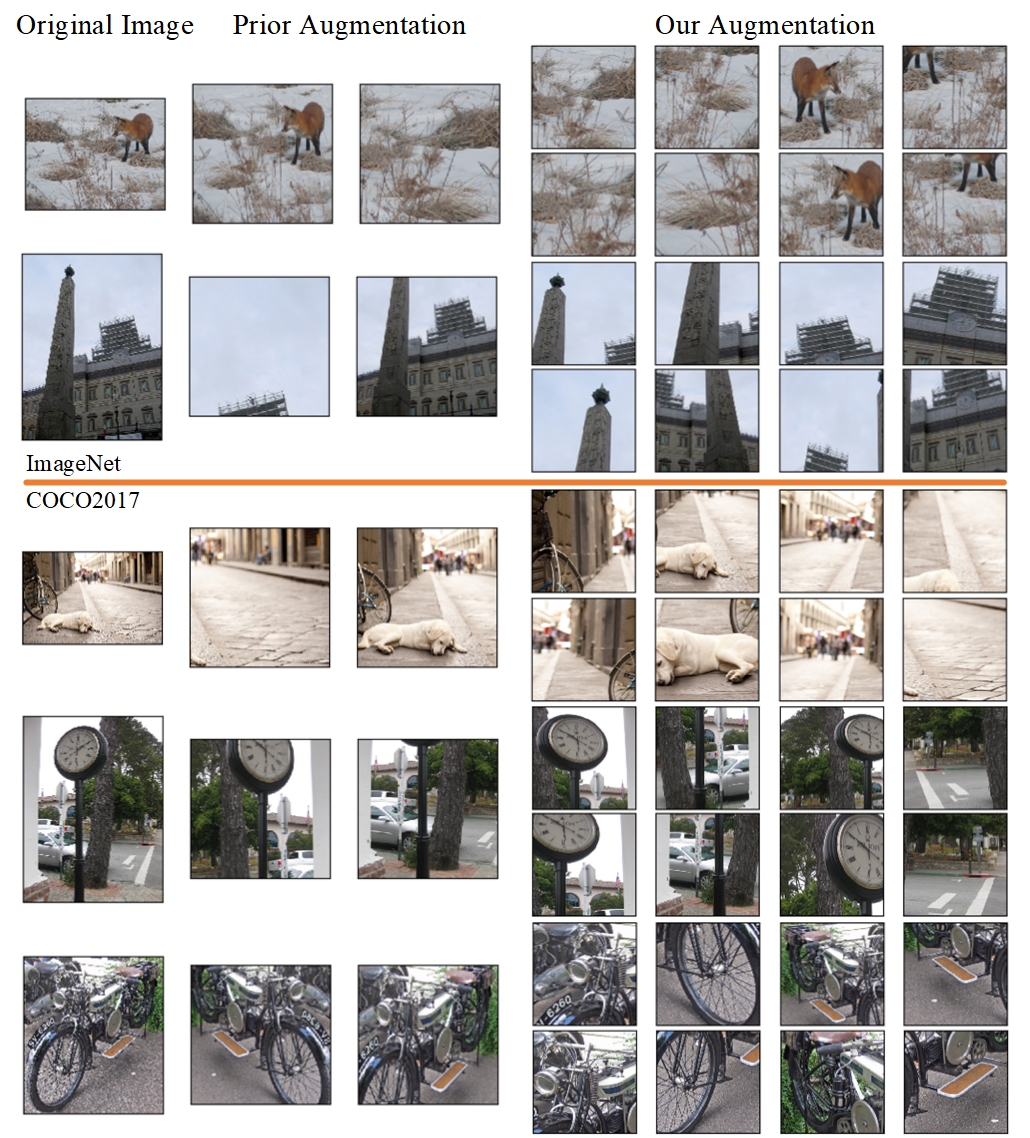}

   \caption{Visual comparison between the conventional positive sample pair generation method (left) and our approach (right) applied to the classical single-label dataset (ImageNet) and the multi-label dataset (COCO2017).}
   \label{fig:example}
\end{figure}

On the other hand, mainstream self-supervised contrastive learning methods  adhere to the similar image augmentation strategy as SimCLR~\cite{simclr}. This strategy involves two random crops on an image, followed by a sequence of other image augmentation operations, ultimately yielding a pair of $224\times224$ views as the actual input for the contrastive learning. While this approach effectively generates two  views as positive sample pairs, it places stringent demands  on dataset quality. In cases where images do not conform to a single target center, it becomes easy to generate erroneous positive view pairs, result in a deviation from the objective of contrastive learning~\cite{rethink,what}. As illustrated  on the left side of ~\cref{fig:example}. When applied to COCO2017, the same image augmentation is more likely to produce two views with entirely distinct semantics, as exemplified by the dog-road and clock-car in the third and fourth rows of ~\cref{fig:example}.  Besides, due to wide-ranging random cropping and resizing, the view generated in each training iteration may contain less than a quarter of the original image. For multi-label images with larger average scales, this  wastes more potential information. In general, randomized view pair generation and limited information utilization, both of which may result in high demands on image quality and quantity.

To mitigate these limitations, we propose a SimSiam~\cite{simsiam}-based SSL pipline specialized for multi-label images. Our approach introduces the Block-wise Augmentation Module (BAM) to  generate  more semantically similar postive view pairs from multi-label images.  Consequently, the multi-block Image-Aware Contrastive Loss (IA-CLoss) is specifically designed to mine the potential semantic relationships among  these views. Both of these innovations  empower our method to achieve  competitive transfer performance, even with fewer and more complex multi-label images. Additionally, we enhance the previous fine-tuned linear validation method for multi-label scenarios. This refinement not only reduces reliance on single-label data but also expands the potential application of self-supervised learning across various scenarios. Our contributions can be summarized as follows:
\begin{enumerate}
\item We extend the SimSiam-based SSL framework to  excel with more general yet complex  multi-label images, which are surprisingly neglected.

\item We propose a block-wise argumentation module and an image-aware contrastive loss  to enable exploitation of more potentially semantically consistent representations in  multi-label images.

\item Without relying on extensive amounts of single-labeled images, our approach achieves competitive transfer learning performance in pre-training with more challenging multi-label datasets.

\end{enumerate}

\section{Related Work}
\label{sec:related}

\subsection{Self Supervised Learning}
In the pursuit of acquiring generalized discriminative features without manual annotation,  early self-supervised learning methods~\cite{prediction1,rotations,jigsaw,mask} struggled to learn semantic consistency representations in solving simple tasks as humans. However, they rarely obtained satisfactory results compared to supervised learning.

The landscape changed with the emergence of the MoCo~\cite{moco1,moco2} and SimCLR~\cite{simclr,simclr2} series, which demonstrated that self-supervised learning could  achieve performance comparable to supervised learning on the ImageNet. This has motivated numerous subsequent studies related to self-supervised learning to shift their focus to contrastive learning~\cite{BYOL,swav,Embedding}. However, these methods tended to pre-training on high-sample datasets, resulting  in unbearable time and resource overheads.  Hence, studies explored  SSL with smaller but task-aligned datasets.  Specifically, some researchers~\cite{does,benchmarking,rethink} employed   naturalistic datasets with fewer samples but finer-grained categories for pre-training to tackle fine-grained classification and image understanding. GeoSSL~\cite{remote} highlighted the value of remotely sensed imagery datasets for pre-training, while   Alina~\etal\cite{ship}  merged multiple remote sensing image datasets, outperforming ImageNet pre-training for specific tasks. Similarly, in  agriculture, Lin~\etal\cite{leaf} relied only on a limited crop-related dataset for pre-training and achieved transfer learning success in leaf segmentation.

\begin{figure*}
  \centering
  \includegraphics[width=1\linewidth]{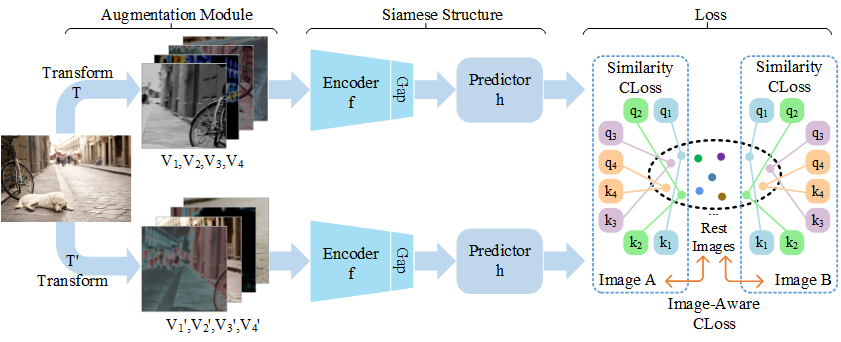}
  \caption{The overall framework of our SSL pre-training method. Firstly,  the image is partitioned  into four regions, each  undergoing two image transformation to generate positive view pairs. Subsequently, all these views enter the twin encoder and predictor, yielding embedding vectors  $q_i$ and $k_i$ for each view. Finally, similarity contrastive loss (CLoss) is calculated between pairs of views from the same region. Simultaneously, all views from the same image are treated as positive samples for the image-aware CLoss.}
  \label{fig:framework}
\end{figure*}

In this paper, our focus will be on contrastive learning approaches, aiming to attain competitive self-supervised learning performance with fewer and more complex multi-label image inputs.

\subsection{Contrastive Learning}
Contrastive learning (CL) is a method to discover invariant image representations by comparing artificially constructed pairs of positive and negative samples~\cite{survey4,survey5,survey6,survey7}. Its simple architecture and firm  performance in SSL have led to increased research interest.

In earlier studies, MoCo~\cite{moco1} used a memory bank to store encodings of previous samples for creating negative samples. At the same time,  SimCLR~\cite{simclr} extended the use of image augmentation to generate positive sample pairs, achieving results similar to supervised learning. The usage of multiple image augmentation methods and the introduction of the multilayer perceptron (MLP) header have also triggered a series of subsequent discussions and research~\cite{moco2,simclr2,MLP}. Improvingly, BYOL~\cite{BYOL}, SwAV~\cite{swav} ,and SimSiam~\cite{simsiam} integrated clustering and prediction processes, which not only streamline the model but also enhance  performance by eliminating negative samples.

Inspired by the studies above, there has been a growing interest in advancing finer-grained comparisons.  Some researchers~\cite{CLAST,CoDo} integrated  background information from downstream scenes during pre-training, improving  robustness.  Similarly, Xu~\etal\cite{Seed} adopted cutmix-like image augmentation during pre-training, expanding the view of a single image to encompass other related samples, thereby enhancing the acquisition of consistent semantics. Further, Zhao~\etal\cite{Embedding,Global,selection,LESSL} combined local views from different images as inputs to the encoder and subsequently decoupled them in the feature dimensions to generate additional sample pairs.   DenseCL~\cite{Dense} adopted a granular approach, constructing contrast pairs among pixels in encoded features. 

However, these studies are almost exclusively pre-trained on single-label datasets, and the data requirements may contradict the original intention of self-supervised learning to achieve wide applicability~\cite{what}. This prompted us to explore the potential of multi-label images, which are more accessible and widely employed in various downstream tasks.

\section{Method}
\label{sec:method}
In our proposed method, we share the same overall architecture with SimSiam~\cite{simsiam}, which has gained popularity in numerous  studies due  of its simplicity  and  freedom from  negative samples. ~\cref{fig:framework} provides an overview of our pre-training network, which encompasses three crucial components: augmentation module, twinned encoding and perception network, and the computation of losses.

\begin{figure*}
  \centering
  \includegraphics[width=1\linewidth]{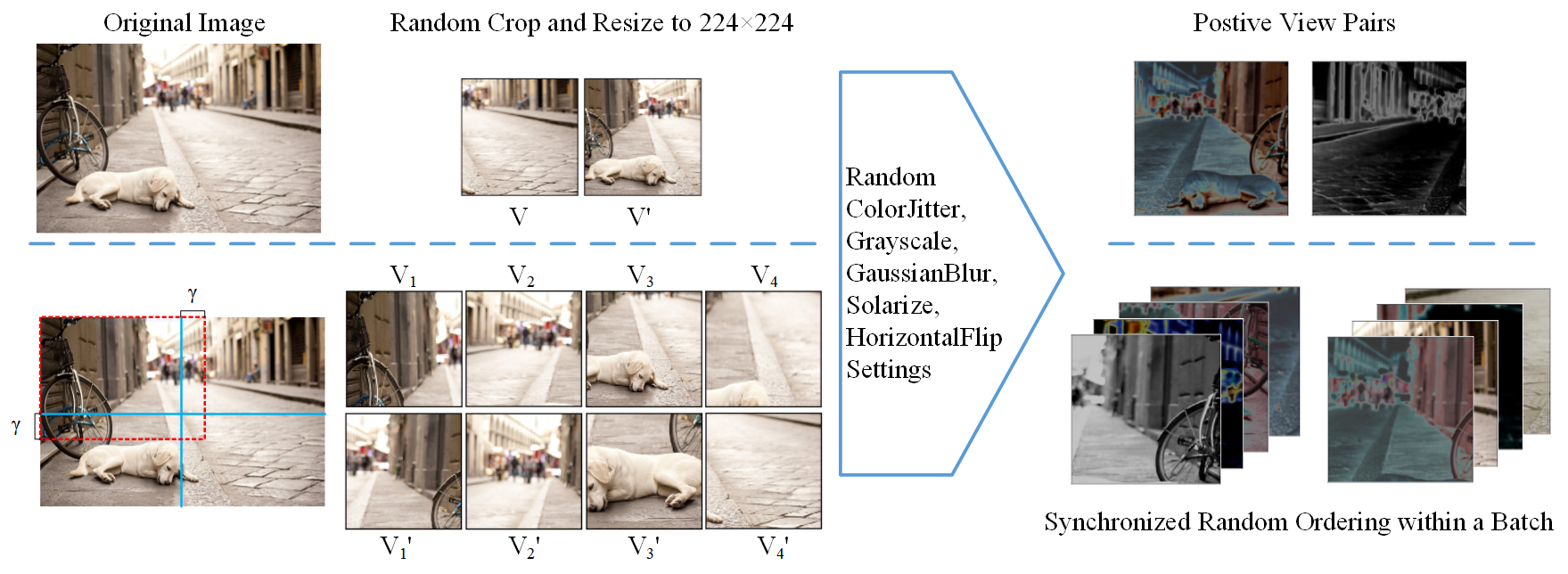}
  \caption{Comparison between the prevalent methods for constructing positive view pairs in contrastive learning and our block-wise augmentation. 
}
  \label{fig:bam}
\end{figure*}

To expand the self-supervised contrastive learning to multi-label images, we introduce our improvements while keeping the main framework of contrastive learning intact. First, we propose the Block-wise Augmentation Module  to generate more  semantically similar pairs of positive views. Further, considering  the potential semantic connections between these  views, we design the multi-block Image-Aware Contrastive Loss (IA-CLoss) based on supervised contrastive loss ~\cite{scl}. Additionally, the previous linear evaluation strategy is improved to make it suitable for multi-label inputs.  In the forthcoming sections, we will provide a comprehensive workflow and a thorough analysis of the motivation behind each component.

\subsection{Block-wise Augmentation Module (BAM)}\label{subsectionbam}
To enhance the extraction of valuable information from multi-label images, we propose a block-wise augmentation module  to generate a more extensive set of  reasonable positive view pairs. This module divides the original image into four overlapping blocks of equal size before  subsequent image augmentation families. The schematic representation of this division is illustrated by the red dashed box in ~\cref{fig:framework}. Each block is scaled to $50\% + \gamma$ of the original image, where $\gamma$ is a manually specified number within the range of $0\%$ to $50\%$, and the default is $20\%$. Subsequently, independent image augmentation operations, such as  color jittering, grayscale conversion,  and flipping, are performed on each  block,  typically employed by mainstream contrastive learning methods. Finally, to prevent  an unintentionally biased distribution where views of the four blocks within the same sample consistently appear adjacent, we specifically obfuscate the order of all pairs obtained within a batch.

Notably,  to take into account the typically larger scale of multi-label images and  positive view pair mis-matching caused by previous image augmentation. Our approach incorporates random cropping and resizing within each block with the standard dimensions of $224\times224$. This methodology  is similar to, but distinct from, the multi-crop operation in SwAV~\cite{swav}, which generates more small-scale views to mine information about local details. Our approach aims to decentralize the past limitation of treating the entire image as a single semantic entity~\cite{simclr,moco1,dino}  and discover possible different semantic targets in the multi-label image. In this way, we not only fully utilize the original image information, but the smaller blocked regions are also more likely to generate visually similar positive view pairs, effectively minimize mismatches.   As depicted in ~\cref{fig:example}, compared with prior augmentation, the view pairs generated by  individual sub-blocks exhibit greater visual and semantic similarity, capturing nearly all potential information present in the original image. Furthermore, in scenarios  where the image contains only a single primary target, as exemplified in the fifth row of ~\cref{fig:example}, the generation of view pairs within  each block enhances the scrutiny of details within each localized region.

As shown in ~\cref{fig:loss,tab:transform}, the incorporation of BAM  not only greatly expedites the convergence of the similarity loss branch but also yields comparable transfer learning performance despite discrepancies in data quantity and quality. Surprisingly, our concern that too  similarity between view pairs would cause the model collapsing or overfitting did not occur, even with a batch size of 128. And as with most self-supervised methods significantly benefit from a larger batch size. We contend that these enhancements can be attributed to the inherent complexity of multi-label images, which exhibit larger image size and the co-occurrence of multiple targets, as well as the untapped potential of contrastive learning.

\subsection{Image-Aware Contrastive Loss (IA-CLoss)}
Intuitively, several targets appearing on the same image often exhibit inherent semantic connections, such as the associations between a person and a bicycle or a plate and a fork. This co-occurrence correlation has been widely explored and employed to enhance the capacity of representation learning in mainstream multi-label image classification methods ~\cite{gcn,stmg,co,cpcl,att}. Moreover, our block-wise augmentation process may fragment a uniform target in the original image, as exemplified in row 5 of ~\cref{fig:example}, where a complete motorcycle is divided into four localized views that exhibit reduced visual similarity. To achieve the semantically consistent representations necessary for motorcycle recognition, establishing connections between these individual views is necessary.

Considering the above insights, we propose the multi-block Image-Aware Contrastive Loss, denoted as $L_{ia}$. $L_{ia}$ minimizes the embedding distance between views from the same image and those from other images, guided by precise supervision signals. These signals are derived from the BAM, as discussed in ~\cref{subsectionbam}. Importantly, due to the block-wise augmentation and the siamese architecture, each original sample yields eight views. We anticipate that these views will be closer  to each other than those from different samples, as they originate from the same image. $L_{ia}$ can be formalized using the following equation:
\begin{equation}\label{eq1}
L_{ia} = -\sum_{i\in I} \frac{1}{\left| P(i) \right|} \log \frac{\sum_{p\in P(i)} \exp \left( \frac{z_i \cdot z_p}{\tau} \right)}{\sum_{a\in A(i)} \exp \left( \frac{z_i \cdot z_a}{\tau} \right)},
\end{equation}
where $\cdot$ represents the inner product of the embedding vectors, and $\tau$ is a temperature parameter.  Here, $A(i)$ signifies the set of all views generated in the batch except for view $i$. Subsequently, $P(i)$ is a subset of $A(i)$, including only  views from the same image as sample $i$.

The structure of the IA-CLoss bears a resemblance to  the supervised contrastive loss~\cite{scl}, which encourages multiple views of the same category to be more proximate through the utilization of category annotations in the contrastive learning of single-label images. Nevertheless, when the input samples are multi-labeled images encompassing multiple objectives, the efficacy of supervised contrastive loss becomes challenging. Since the partitioned views may exhibit substantial semantic disparities, as illustrated in row 3 of ~\cref{fig:example}.

Using sample $i$ as an illustration, we compute and simplify its gradient, resulting in the form presented in ~\cref{eq2}. The magnitude of the final gradient is contingent upon the mean of the gaps between the embedding vectors in $P(i)$ and $A(i)-P(i)$. Different from supervised contrastive loss, IA-CLoss exhibits a gentler behavior that aligns better with the requirements of contrastive learning in the context of multi-label inputs. This is substantiated by the results depicted in ~\cref{tab:t224,tab:t448}, where IA-CLoss consistently demonstrates an ongoing enhancement in model performance over extended iterations, in contrast to the diminishing performance observed with supervised contrastive loss (SUP).

\begin{align}\label{eq2}
&\frac{\partial L_{ia}}{\partial z_i} = -\frac{1}{|P(i)|\tau} \frac{\sum_{p \in P(i)}{z_p e^{\frac{z_i \cdot z_p}{\tau}}} - \sum_{a \in A(i)}{z_a e^{\frac{z_i \cdot z_a}{\tau}}}}{\sum_{p \in P(i)}{e^{\frac{z_i \cdot z_p}{\tau}}} \sum_{a \in A(i)}{e^{\frac{z_i \cdot z_a}{\tau}}}} \\
&= -\frac{1}{|P(i)|\tau} \frac{\sum_{p \in P(i),a \in A(i) - P(i)}{(z_p - z_a)e^{\frac{z_i \cdot z_p}{\tau}}e^{\frac{z_i \cdot z_a}{\tau}}}}{\sum_{p \in P(i),a \in A(i)}{e^{\frac{z_i \cdot z_p}{\tau}}e^{\frac{z_i \cdot z_a}{\tau}}}}\nonumber
\end{align}

The total loss for pretraining can be formulated as follows:
\begin{equation}\label{eq3}
L_{total}=L_{sim}+\lambda \cdot L_{ia},
\end{equation}
where $L_{sim}$ is the cosine  similarity contrastive loss~\cite{simsiam} and $\lambda$ denotes the weight parameters of $L_{ia}$, which has a default value of 4. 

\begin{table*}
  \centering
  \begin{tabular}{@{}lcccccccccccc@{}}
    \toprule
    Method &Batch Size &Epoch & BAM & IA-CLoss &  GAMP & mAP & OP & OR & OF1 & CP & CR & CF1\\
    \midrule
    Sim &512&200 &  &  &   & 35.5 & 86.6 & 17.9 & 29.7 & 32.1 & 5.7 & 9.7 \\
	Sim\textsuperscript{IN} &512&100 &  &  &   & 41.1 & 93.4 & 13.9 & 24.2 & 19.5 & 1.3 & 2.5\\
    Our &128& 100 &\checkmark &  &   & 45.0 & 91.5 & 13.5 & 23.5 & 9.5 & 1.1 & 2.0 \\
    Our & 128&200 &\checkmark &  &   & 47.9 & 88.7 & 11.8 & 20.8 & 1.21 & 0.8 & 1.0\\
	Sim &512& 200 & &  &  \checkmark & 50.9 & 85.9 & 33.9 & 48.6 & 74.0 & 24.6 & 36.9\\
	Our &128& 100 &\checkmark &  &  \checkmark & 54.3	&88.3	&38.0	&53.2	&75.1	&27.2	&40.0\\
	MoCo\textsuperscript{IN} &256&200 &  &  &   & 56.5 & 90.0 & 35.8 & 51.2 & 80.8 & 25.5 & 38.7\\
	Sim\textsuperscript{IN} &512&100 &  &  &  \checkmark & 56.6 & 89.7 & 36.7 & 52.1 & 80.3 & 26.3 & 39.6\\
	Our &128& 200 & \checkmark &  & \checkmark & 57.2 & 88.3 & 38.0 & 53.2 & 75.1 & 27.2 & 40.0\\
	Our & 128&100 &\checkmark & \checkmark &  \checkmark & 60.2 & 87.7 & 42.5 & 57.2 & 77.5 & 33.9 & 47.1\\
	Our &128& 200 &\checkmark & SUP &  \checkmark & 62.0 & 88.0 & 43.1 & 57.9 & 80.0 & 35.6 & 49.3\\
	Our &128& 100 &\checkmark & SUP &  \checkmark & 62.3 & 88.2 & 43.8 & 58.5 & 81.4 & 35.7 & 49.7\\
	MoCo\textsuperscript{IN} &256&200 &  &  &  \checkmark & 63.3 & 88.6 & 44.5 & 59.3 & 83.0 & 37.0 & 51.2\\
	Our & 128&200 &\checkmark & \checkmark &  \checkmark & 64.0 & 88.6 & 45.6 & 60.2 & 83.1 & 38.3 & 52.4\\
    \bottomrule
  \end{tabular}
  \caption{Comparison of our approach with  SimSiam (Sim) and MoCoV2 (MoCo) in the fine-tuned multi-label classification on COCO2017, with an input scale of $224\times224$. Sim\textsuperscript{IN} and MoCo\textsuperscript{IN} are weights pre-trained on ImageNet; SUP uses a supervised contrastive loss instead of our IA-CLoss; GAMP records whether the fine-tuning and validation phases introduced  global max pooling.}
  \label{tab:t224}
\end{table*}

\begin{table*}
  \centering
  \begin{tabular}{@{}lcccccccccccc@{}}
    \toprule
    Method &Batch Size &Epoch & BAM & IA-CLoss &  GAMP & mAP & OP & OR & OF1 & CP & CR & CF1\\
    \midrule
	Sim\textsuperscript{IN} &512&100 &  &  &   & 38.8 & 94.3 & 11.3 & 20.1 & 2.5 & 0.8 & 1.2\\
    Sim &512&200 &  &  &   &42.2 & 85.6 & 27.1 & 41.2 & 59.7 & 14.5 & 23.4 \\
    Our &128& 100 &\checkmark &  &   & 44.3 & 89.3 & 12.0 & 21.2 & 2.2 & 0.8 & 1.2 \\
    Our & 128&200 &\checkmark &  &   & 47.3 & 93.6 & 10.3 & 18.6 & 1.2 & 0.7 & 0.9\\
	Sim &512& 200 & &  &  \checkmark & 52.3 & 86.4 & 35.8 & 50.7 & 71.6 & 25.6 & 37.7\\
	MoCo\textsuperscript{IN} &256&200 &  &  &  & 56.0 & 91.1 & 34.0 & 49.5 & 79.0 & 23.5 & 36.2\\
	Our & 128&100 &\checkmark &  &  \checkmark & 58.8 & 89.0 & 39.8 & 55.0 & 74.1 & 29.0 & 41.7\\
	Sim\textsuperscript{IN} &512&100 &  &  &  \checkmark & 61.1 & 91.1 & 39.3 & 54.9 & 76.2 & 28.7 & 41.7\\
	Our & 128&200 & \checkmark &  & \checkmark & 63.3 & 91.0 & 42.4 & 57.8 & 82.2 &32.0 & 46.1\\
	Our &128& 100 &\checkmark & \checkmark &  \checkmark & 66.2 & 90.0 & 47.6 & 62.3 & 84.8 & 40.5 & 54.8\\
	Our &128& 200 &\checkmark & SUP &  \checkmark & 66.7 & 90.7 & 48.0 & 62.8 & 86.0 & 40.6 & 55.2\\
	Our &128& 100 &\checkmark & SUP &  \checkmark & 67.9 & 91.0 & 48.1 & 62.9 & 85.0 & 41.6 & 55.8\\
	MoCo\textsuperscript{IN} &256&200 &  &  &  \checkmark & 68.1 & 90.0 & 50.2 & 64.4 & 83.9 & 43.6 & 57.4\\
	
	Our & 128&200 &\checkmark & \checkmark &  \checkmark & 68.5 & 90.1 & 50.6 & 64.8 & 83.7 & 43.7 & 57.4\\
    \bottomrule
  \end{tabular}
  \caption{Comparison of our approach with  SimSiam (Sim) and MoCoV2 (MoCo) in the fine-tuned multi-label classification on COCO2017, with an input scale of $448\times448$.}
  \label{tab:t448}
\end{table*}

\subsection{Improved Fine-tuning Validation Framework for Multi-label Image Inputs}\label{subsectionfine}
A comprehensive review of the fine-tuning validation process for established self-supervised contrastive learning methods reveals a well-recognized practice:  initial  pre-training on ImageNet, followed by weight freezing for training single-label classification heads during the validation. Nevertheless, its applicability to multi-label inputs presents a formidable challenge. As indicated in the rows without GAMP in ~\cref{tab:t224}, we replaced only the single-label classifiers and losses in the traditional fine-tuned validation framework with their multi-label counterparts. The pre-training weights in almost all conditions struggle to achieve acceptable multilabel prediction performance. Even the top-performing MoCo\textsuperscript{IN}~\cite{moco2} suffers from  significant accuracy and recall.   Despite the utilization of larger inputs ($448\times448$) in the fine-tuning training and validation process, it only produces a modest enhancement in both accuracy and recall, remaining considerably below acceptable thresholds (see ~\cref{tab:t448}).

The high accuracy but low recall in the prediction results suggests that while most of the identified labels in the image are correct, many  labels are still overlooked. We argue that the global average pooling applied at the end of the encoder during pre-training and fine-tuning diminishes the spatial structural information within the image ~\cite{shuffled,csra,mcar}. Additionally, the contrastive loss, post multilayer perceptron mapping,  treats the entire view as a single category for representation learning \cite{MLP,does,simclr}. Consequently, achieving acceptable multi-label prediction results with frozen encoder weights remains  challenging, even when the single-label classifier and loss function are replaced with their multi-label counterparts during fine-tuning.

Based on the aforementioned experiments and analysis, we sought to validate the efficacy of pre-trained weights with a straightforward linear classifier, even when working with multi-label datasets. To this end, we propose to replace the last global average pooling  of the original encoder as the mean of global average pooling minus global max pooling (GAMP). This adjustment enables  the same pre-trained weights to  effectively adapt to multi-label image classification with minimal additional parameters or complex structural modifications to the classifier. As illustrated in ~\cref{tab:t224} and ~\cref{tab:t448}, the integration of global max pooling  results in significant improvements in the fine-tuned validation metrics, despite the recall remaining lower than the accuracy.


In summary, we have achieved targeted optimization of multi-label images in the self-supervised learning. This allows us to extract valuable information from a smaller and more complex dataset. Furthermore, we have improved the traditional linear fine-tuning process, enabling quick validation with multi-label annotations. These  advancements will expand the use of self-supervised learning to a wider range of situations where single-label images or annotations are not easily accessible.

\section{Experiment}
\label{sec:experiment}

\subsection{Experimental Details}

\textbf{For  pre-training:} our approach  utilizes the training set of the classical multi-label image dataset, COCO2017~\cite{coco}, comprising  $117,266$ images. Compared with ImageNet, COCO2017 exhibits a larger average image scale ($577\times484$ vs.  $472\times405$). Furthermore, due to the  coexistence of multiple targets in COCO2017,  self-supervised learning becomes significantly more challenging.

For training parameter settings, we use Resnet50 as the encoder and follow almost all of SimSiam~\cite{simsiam}.   The learning rate is initially set at $0.05\times \text{BatchSize}/256$ with cosine decay.  For a fairer  comparison, the methods incorporating  BAM employs  a batch size of 128 and maintain a consistent initial learning rate of 512, adjusting for the fourfold increase in the number of view pairs introduced by BAM.

\begin{table*}
  \centering
\resizebox{\textwidth}{!}{
  \begin{tabular}{@{}lcccccccccccccc@{}}\toprule

\multirow[m]{2}[2]{*}{\textbf{Method}}	&\multirow[m]{2}[2]{*}{\textbf{BAM}}	&\textbf{IA}	&\textbf{Pre} &\multirow[m]{2}[2]{*}{\textbf{Epoch}}&\textbf{Batch}& \multicolumn{3}{c}{\textbf{VOC BBOX}} &\multicolumn{3}{c}{ \textbf{COCO BBOX}}&\multicolumn{3}{c}{ \textbf{COCO SEG}}\\\cmidrule(lr){7-9}\cmidrule(lr){10-12}\cmidrule(lr){13-15}
&&\textbf{CLoss}&\textbf{Dataset}&&\textbf{Size}& \textbf{AP}	& \textbf{AP50}& \textbf{AP75}& \textbf{AP}& \textbf{AP50}& \textbf{AP75}& \textbf{AP}& \textbf{AP50}& \textbf{AP75}\\

    \midrule
     SimSiam&&& ImageNet &100 & 512 & 38.2  & 66.0 & 38.2 & 29.8 & 47.9 & 31.3 & 26.9 & 45.1 &28.3\\
    Our&\checkmark && COCO&200&128& 36.5 & 64.4  & 35.8 & 29.9 & 48.0 & 31.6 & 27.0 & 45.3 & 28.4\\
	Our&\checkmark &\checkmark&COCO&200&128& 46.2 &75.7 & 48.7 & 31.6 & 50.5 & 33.7 & 28.4 & 47.6 & 29.8 \\
	MoCoV2&&&ImageNet&200&256&45.7  & 74.1 & 48.7 & 31.9 & 50.5 & 34.1 & 28.5 & 47.5 &29.9\\
DenseCL&&&COCO&800&256&47.4  & 76.2 & 50.2 & 32.2 & 50.9 & 34.0 & 28.7 & 47.8 &30.2\\
Supervised&&&ImageNet&-&-&47.9  & 77.0 & 50.8 & 32.9 & 52.0 & 35.1 & 29.3 & 48.8 &30.8\\
	Our&\checkmark &\checkmark&COCO&200&256& \pmb{49.7} & \pmb{79.0} & \pmb{53.5} & \pmb{33.6} & \pmb{53.0} & \pmb{35.8} & \pmb{29.9} & \pmb{50.0} & \pmb{31.2} \\
    \bottomrule
  \end{tabular}}
  \caption{Performance of different pre-training weights on VOC object detection (VOC BBOX), COCO object detection (COCO BBOX) and COCO semantic segmentation (COCO SEG). Order by AP in COCO BBOX.}
  \label{tab:transform}
\end{table*}

\textbf{For the liner evaluation protocol:} to cope with multi-label inputs, we propose the fine-tuned verification framework in ~\cref{subsectionfine}. The specific parameter settings are as follows: 100 epochs of iterative training with an initial learning rate of $10\times BatchSize/256$, a cosine decay schedule for  learning rate updates, and the use of SGD as the optimizer with weight decay (0.0001) and  momentum (0.9).

Furthermore, additional evaluation metrics are employed for multi-label recognition. These metrics include the average precision (AP) for each category and the mean average precision (mAP) across all categories, which serve as the primary measures. To facilitate comprehensive comparisons, we also report class-wise precision (CP), recall (CR), F1 score (CF1), and overall precision (OP), recall (OR), and F1 score (OF1).

\textbf{For  object detection and segmentation:} we  adopt the same transfer training strategy as SimSiam~\cite{simsiam}, leveraging the publicly available MoCo codebase~\cite{moco1,moco2} to assess the effectiveness of pre-trained weights for 1x scheduler object detection and  segmentation on COCO2017~\cite{coco}, as well as 24k object detection on VOC~\cite{voc}. Due to hardware constraints, we reduce the batch size to 8. In ~\cref{tab:transform}, we document the batch size, epoch, and dataset used for each weight pre-training, and ensure consistency across all remaining transfer training configurations. Notably,  all pre-trained weights on ImageNet are sourced from publicly available weights provided by SimSiam~\cite{simsiam}, MoCoV2~\cite{moco2}, and DenseCL~\cite{Dense}.

\begin{figure}[t]
  \centering
   \includegraphics[width=1\linewidth]{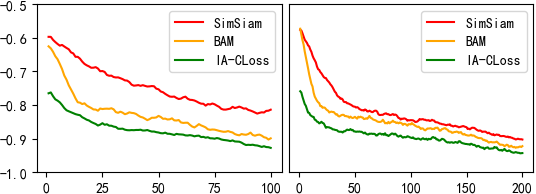}

   \caption{Visualization of trends in similarity contrastive loss after sequential introduction of BAM and IA-CLoss at  pre-training on COCO2017. The horizontal axis indicates pre-trained epochs.}
   \label{fig:loss}
\end{figure}

\subsection{Experimental Results}
In this section, we will provide a comprehensive description and analysis of the experimental results obtained in our study.

\textbf{For pre-training:}   as can be observed in ~\cref{fig:loss}, our image augmentation method (BAM) significantly accelerates the convergence speed of the similarity contrastive loss  compared to the baseline (SimSiam), particularly with a reduced  number of iterations. This benefited from BAM mining more  valuable view pairs.

In addition, the introduction of IA-CLoss  based on BAM further assists  the convergence of the similarity contrastive loss (~\cref{fig:loss}).  The fine-tuning validation results, in conjunction with ~\cref{tab:t224,tab:t448}, demonstrate that the introduction of IA-CLoss performs notably better with extended iteration epochs, while  SUP struggles to bring about further enhancements.

Combined with the transfer performance in ~\cref{tab:transform}, the introduction of BAM significantly increases the quantity and quality of positive view pairs,  even achieving  comparable results to ImageNet pre-training in downstream tasks at the same domain. Furthermore, when dealing with the co-occurrence of multiple targets in multi-label images, IA-CLoss effectively establishes inter-view connections, resulting in a more coherent and consistent semantic representation. This enhanced representation aligns more closely with the requirements of dense prediction.

\textbf{For linear fine-tuning validation in multi-label image classification:} ~\cref{tab:t224,tab:t448} demonstrate  that the simple addition of global max pooling to the last layer of the  encoder during fine-tuning training and validation can improve recall without compromising accuracy. This corroborates the property of contrastive learning, where the entire view is treated as a category, and the unique advantage of linear validation on single labels~\cite{what}. Additionally, the global pooling at the end of the encoder smooths out spatial information, which holds significance in multi-label classification~\cite{shuffled,csra}. 

More fundamentally, this discovery bridges the gap of previous SSL methods for fine-tuned validation on multi-label  datasets. This  further frees self-supervised learning from its reliance on large-scale single-label datasets, enabling its adaptability to a wider range of scenarios.

\textbf{For transfer learning:} as shown in ~\cref{tab:transform}, after the introduction of BAM, the COCO pre-trained model exhibited superior performance compared to ImageNet in certain aspects, indicating that more valuable information was extracted from the multi-label input. Furthermore, the reduced domain shift resulted in a significantly improved transfer performance on COCO compared to VOC. With the incorporation of our IA-CLoss branch, all indicators exhibit a substantial improvement, indicating the extraction of semantic representations that better align with the requirements of object detection and semantic segmentation. In contrast to MoCoV2~\cite{moco2} and DenseCL~\cite{Dense}, which are trained with larger datasets or more epochs, our approach maintains a competitive edge in some metrics. In experiments where the batch size is increased, our method stably improves and even surpasses the performance of supervised pre-training. This result further validates the stability and robustness of our method.

\begin{table}
  \centering
  \begin{tabular}{@{}lccccccc@{}}
    \toprule
    $\gamma$  & mAP & OP & OR & OF1 & CP & CR & CF1 \\
    \midrule
    0\%  & 45.6 &84.5	&28.8	&43.0	&63.8	&16.5	&26.3 \\
    10\% & 51.0	&85.7	&33.1	&47.8	&68.1	&21.5	&32.6 \\
    20\% & 54.3	&88.3	&\pmb{38.0}	&\pmb{53.2}	&\pmb{75.1}	&\pmb{27.2}	&\pmb{40.0} \\
	30\% & \pmb{55.5}	&\pmb{88.6}	&36.6	&51.8	&74.0	&25.8	&38.3 \\
	40\% & 54.7	&\pmb{88.6}	&36.0	&51.2	&71.7	&25.0	&37.0 \\
    \bottomrule
  \end{tabular}
  \caption{Performance under different overlap rates ($\gamma$) between blocks in BAM.}
  \label{tab:tau}
\end{table}

\subsection{Ablation Study}
In this section, we conduct ablation studies by self-supervised training of 100 epochs on COCO2017 and quickly validate the pre-training effects through the multi-label fine-tuned validation framework in ~\cref{subsectionfine}. 

Firstly, different values of block overlap rate $\gamma$ in the BAM at $[0\%, 10\%, 20\%, 30\%, 40\%]$ are tested without IA-CLoss. ~\cref{tab:tau} indicates that relatively large overlap rates ($30\%$, $40\%$) have better mAP and OP performance, while  the best balance of accuracy and recall (CF1, OF1) is achieved at $20\%$. This suggests that cropping with too small overlap rates may disrupt targets along the dividing line of the original image, missing the opportunity to encompass the entire target within the view.  Conversely, the reasonableness of positive view pairs is hard to guarantee. In addition, all the blocking methods except no overlap achieve good performance, proving the significant advantage of BAM in dealing with multi-label image inputs.

\begin{table}
  \centering
  \begin{tabular}{@{}lccccccc@{}}
    \toprule
    Zoom  & mAP & OP & OR & OF1 & CP & CR & CF1 \\
    \midrule
    0.1-1.0 & 53.3	&87.1	&34.7	&49.6	&69.0	&23.3	&34.8 \\
    0.2-1.0 & \pmb{54.3}	&\pmb{88.3}	&\pmb{38.0}	&\pmb{53.2}	&\pmb{75.1}	&\pmb{27.2}	&\pmb{40.0} \\
    0.4-1.0 & 52.1	&87.0	&34.0	&48.8	&70.5	&22.4	&33.9 \\
	0.6-1.0 & 39.6	&84.7	&23.3	&36.6	&48.0	&11.1	&17.9 \\
    \bottomrule
  \end{tabular}
  \caption{Performance of our method in randomized cropping with different zoom ratio.}
  \label{tab:deflation}
\end{table}

On the other hand, while keeping $\gamma$ at $20\%$, we investigate various lower bound values for the zoom ratio in random cropping during image augmentation (default $0.2-1.0$), considering values of $[0.1, 0.2, 0.4, 0.6]$. Combined with ~\cref{tab:deflation}, it is intriguing to observe that using the same set of zoom scales as the mainstream method yields optimal performance within BAM. This finding further emphasizes the importance of fully utilizing the valuable information in the original image.

\section{Conclusion}
\label{sec:conclusion}

In this paper, we highlight the reliance of previous self-supervised contrastive learning approaches on single-label datasets, as well as the limitations of conventional positive view pair construction methods. To address these problems, we extend the SimSiam framework to propose block-wise argumentation and image-aware contrastive loss. Our approach encourages the  generation of more consistent  pairs of views and discovers  potential correlations among these views, thereby facilitating semantically consistent representation learning. Extensive experiments demonstrate that our approach achieves  effective performances even under multi-label datasets with fewer samples. Furthermore, we enhance the mainstream fine-tuning validation to accommodate multi-label inputs, thereby reducing dependence on single-label data.

\textbf{Limitations.} Our enhancements and experiments primarily focus on SimSiam due to its simplicity and efficiency in architecture, without extensively considering the specificities of other  frameworks such as SwAV~\cite{swav} and  DINO~\cite{dino} \etal. Further experiments are necessary to establish the applicability of these SSL frameworks to multi-label image inputs. Additionally, while the inclusion of global max pooling to accommodate multi-label classification during fine-tuning is practical, it remains insufficient. More practical refinements are needed to address this concern. Furthermore, investigating the disparity between the representations obtained through self-supervised learning in the SimSiam architecture and the representations required for multi-label image classification would be an intriguing avenue for exploration.



{
    \small
    \bibliographystyle{ieeenat_fullname}
    \bibliography{main}
}


\end{document}